\documentclass[conference]{IEEEtran}
\IEEEoverridecommandlockouts
\usepackage{cite}
\usepackage{amsmath,amssymb,amsfonts}
\usepackage{algorithmic}
\usepackage{graphicx}
\usepackage{textcomp}
\usepackage{xcolor}
\usepackage{nicefrac}       % compact symbols for 1/2, etc.
\usepackage{microtype}      % microtypography
\usepackage{lipsum}
\usepackage{graphbox, graphicx}
\usepackage{subcaption}
\usepackage{multirow}
\usepackage{tabularx}
\usepackage{adjustbox}
\usepackage{makecell}
\usepackage{url}
\usepackage{breakurl}

\def\BibTeX{{\rm B\kern-.05em{\sc i\kern-.025em b}\kern-.08em
    T\kern-.1667em\lower.7ex\hbox{E}\kern-.125emX}}
\begin{document}

\title{Vision-Aided Beam Tracking: Explore the Proper Use of Camera Images with Deep Learning}
\author{\IEEEauthorblockN{Yu Tian, Chenwei Wang}
\IEEEauthorblockA{DOCOMO Innovations Inc., Palo Alto, CA 94304}
Email: \{yu.tian, cwang\}@docomoinnovations.com}

\maketitle

\begin{abstract}
We investigate the problem of wireless beam tracking on mmWave bands with the assistance of camera images. In particular, based on the user's beam indices used and camera images taken in the trajectory, we predict the optimal beam indices in the next few time spots. To resolve this problem, we first reformulate the ``ViWi" dataset in \cite{icc2020} to get rid of the image repetition problem. Then we develop a deep learning approach and investigate various model components to achieve the best performance. Finally, we explore whether, when, and how to use the image for better beam prediction. To answer this question, we split the dataset into three clusters -- (LOS, light NLOS, serious NLOS)-like -- based on the standard deviation of the beam sequence. With experiments we demonstrate that using the image indeed helps beam tracking especially when the user is in serious NLOS, and the solution relies on carefully-designed dataset for training a model. Generally speaking, including NLOS-like data for training a model does not benefit beam tracking of the user in LOS, but including light NLOS-like data for training a model benefits beam tracking of the user in serious NLOS. 
\end{abstract}

% \begin{IEEEkeywords}
% component, formatting, style, styling, insert
% \end{IEEEkeywords}

\section{Introduction}\label{sec:intro}

Geared towards supporting substantial latency reduction and massive connectivity, the mmWave deployment is expected to be a key in 5G/B5G. Compared to their sub-6GHz counterparts, the channel on mmWave bands change much faster due to higher Doppler, thereby making accurate and timely channel acquisition challenging in applications, such as channel inference, user localization, and beam tracking. Contrary to the conventional beam tracking that relies on the pilot-based channel estimation within a coherent block, machine learning has been attractive to perform proactive beam prediction. While this approach can deal with the users in the line-of-sight (LOS) scenarios, it will likely struggle in real wireless environments with multiple users, multiple blockages, and rich dynamics \cite{icc2020}. In response to that, Alrabeiah \emph{et al.} recently developed a dataset named for ``ViWi" \cite{Alrabeiah_icc, icc2020}. The scenario is characterized by two base stations (BSs), static objects, and dynamic objects, including persons, cars, trucks and buses with different speeds. While the BSs are positioned on the opposite sides of the street to cover the entire street, the transmitted signal could be LOS, NLOS, and blocked. Moreover, three RGB cameras are installed in each BS and takes photos of certain part of the street. For each user, each BS aims to proactively predict the optimal beam in the next few time spots. 

The turn to visual data as a supplementary information for beam tracking is motivated by two key factors: (i) the fact that images are rife with information about the environments they are depicting, and (ii) the major strides computer vision has taken in image comprehension with the help of deep learning. \emph{Compared to using the beam only, whether, when, and how to use the image for better beam prediction} are of interest.

%\subsection[Related Work]

The research interests in vision-aided wireless applications are skyrocketing in academia in recent two years. Starting from \cite{Alrabeiah_resnet}, Alrabeiah \emph{et al.} investigated mmWave link blockage prediction and beam prediction. Assuming each image contains only one object, they directly applied classical image classification models for prediction. Later, when they investigated vision-aided mmWave beam tracking in \cite{Alrabeiah_icc} by using the beam sequence only, they also offered a dataset where each image could contain multiple moving objects, but which user is the target is unknown\footnote{In practice, this assumption might be little ill-posed, as without knowing which object in the image is the target, one cannot claim that user is in the covered area of that camera. Admittedly, this assumption reduces the difficulty from the actual situation, and could be a start vision-aided beam tracking.}. Similar to the baseline in \cite{Alrabeiah_icc} but replacing the uni-GRU with bi-LSTM and bi-GRU, \cite{master_thesis} performed $3\%$ (slightly) better. Then in \cite{Alrabeiah_yolo}, Alrabeiah \emph{et al.} investigated signal blockage prediction. In particular, object detection was applied to extract the image features, followed by further feature embedding. Finally, the sequential image feature embedding and beam embedding were alternatively fed to a uni-GRU for supervised learning. Still in 2020, as the data competition task at IEEE ICC, vision-aided beam tracking attracted nine teams \cite{icc2020}. Although the approaches of the participants have not been available in public, the most interesting observation is that even compared to the baseline in \cite{Alrabeiah_icc}, the scores of all the submissions were much worse in the test dataset (for participants only) \cite{icc2020}. This is usually caused by either over-fitting or more likely, in the competition, the different data distribution of unseen test dataset from the training and validation datasets. The most recent work on resolving this competition task was shown by Tian \emph{et al.} in \cite{Alouini} where a new deep learning model based on ResNet and ResNeXt was proposed to embed the image. Leveraging the power of deeper models, they achieved significantly better scores than those in literature. However, it is still not clear whether the performance is significantly improved by the use of powerful deeper and complicated model for better processing the beam sequence only or the image sequence as well.

% \begin{figure}[t]
%     \centering
%     \includegraphics[align=t, width=1.0\linewidth]{env.png}
%     \vspace{-0.1in}
%     \caption{The environment to produce the ``ViWi" dataset in \cite{Alrabeiah_yolo}}
%     \label{fig:sys}
% \end{figure}

%\subsection{Our Contribution}

In this paper, we investigate the problem of vision-aided beam tracking defined in \cite{icc2020} but with a reformulated new dataset from the ``ViWi" dataset in \cite{icc2020}. More importantly, we shed light on the question compared to using the beam only, whether, when, and how to use the image for better beam prediction. Our contribution can be summarized as follows.

Firstly, in the original ``ViWi" dataset, the training and validation datasets share \emph{nearly all} the images in common. In practice, after a model is trained, the images in the validation dataset should not be seen before even they are taken from the same environment. Thus, for all the prior results in \cite{Alrabeiah_icc, master_thesis, Alouini}, whatever the performance scores have been achieved, it is not clear {\em if the model performs well in validation because it is able to better learn image features or it just memorizes the same image in training}. To get rid of the image repetition in both training and validation, we reformulate a new dataset where the images in training and validation are mutually exclusive.

Second, we develop a deep learning approach based on the extension  of the model used for link blockage prediction in \cite{Alrabeiah_yolo} with several variations. Specifically, in the image feature extraction component we consider CNN, AE, PCA, and in the sequential model component we consider uni-GRU and bi-GRU. Finally, it turns out that the combination of AE and bi-GRU produces the best performance metric value $0.687$.

Finally, during the process of model component selection and parameter tuning, we are aware that compared to using the beam only, the benefit brought by using the image for beam prediction readily becomes marginal when the model becomes deeper and more complicated. Thus, we explore whether, when, and how to use the image for better beam prediction. To answer this question, we split the dataset into three clusters -- (LOS, light NLOS, and serious NLOS)-like -- based on the standard deviation of the beam sequence. Our experiments demonstrate that using the image indeed helps beam tracking especially when the user is in serious NLOS, and the solution relies on carefully-designed dataset for training a model.

\section{Problem Definition}\label{sec:sys}

The problem of vision-aided beam tracking and its evaluation metric were introduced in \cite{icc2020}. For completeness, we briefly re-introduce them in this section and then introduce the newly reformulated dataset from the ViWi dataset in \cite{icc2020}. %, as introduced in Sec. \ref{sec:intro}, to get rid of the image repetition problem in training and validation datasets.

%\subsection{Problem Description and Evaluation Metric}

We assume that each BS is equipped with $N$ antennas operating in 28G Hz band. For each user $u$ at time $t$, a beam ${\bf f}_u(t)$ is selected by the BS from a predefined codebook $\mathcal{F}$ to serve it in the downlink. For proactive beam prediction in each of the next $m$ time spots, the BS can observe the used beams and the corresponding camera images in up to $\tau$ prior time spots. Given a fixed codebook $\mathcal{F}$, since the beam ${\bf f}$ and its index $f$ form a one-to-one mapping, we define an observational sequence for each user instance $u$ as
\begin{eqnarray}
{\bf S}_u(t)\!=\![(f_u(t\!\!-\!\tau\!+\!1),{\bf X}_u(t\!\!-\!\tau\!+\!1)),\cdots,(f_u(t),{\bf X}_u(t))]\!\!\!  
\end{eqnarray}
where ${\bf X}$ is a 3-D tensor representing the camera RGB image. Given an observational sequence ${\bf S}(t)$, we aim to predict the best beams in the next $m$ time instances, denoted by $\hat{f}(t'),~t'=t+1\cdots,t+m$. The ground truth indices for the next $m$ time spots are already given in the dataset, under the sense of maximizing the received SNR at the user. %Thus, our goal in this paper is explore learning-based methods to predict $\hat{f}(t'),~t'=t+1,\cdots,t+m$.

To compare how close the prediction $\hat{f}(t')$ is to $f^*(t')$, we follow the metric proposed in \cite{icc2020}. Specifically, define the score of next $m$ predictions over user instances of interest as
\begin{eqnarray}
\textrm{score}_m = \mathbb{E}_u\mbox{$\left[\exp\left(-\frac{1}{m\sigma}\sum_{t'=t+1}^{t+m}\big|\hat{f}_u(t')-f_u^*(t')\big|\right)\right]$}
\end{eqnarray}
where $\sigma$ is a predefined penalization factor. In Section \ref{sec:modif_ret}, we will leverage this definition and its variations for evaluation. To be consistent with \cite{icc2020} and for comparison with the literature, we explore the metric of $\textrm{TotalScore}$ defined by:
\begin{eqnarray}
\textrm{TotalScore}=  (\textrm{score}_1+3\cdot\textrm{score}_3+5\cdot\textrm{score}_5) / 9.
\end{eqnarray}

\subsection{Dataset Description} \label{sec:dataset}

In this section, we introduce the new training and validation datasets by varying the Viwi dataset \cite{icc2020}. In the Viwi dataset, We denote by $\mathcal{D'}_{t}$ the training dataset that contains $281100$ user instances (rows) and $\mathcal{D'}_{v}$ the validation dataset with $120468$ user instances (rows). Each instance has an observational sequence ${\bf S}(t)$ with length $\tau = 8$ and the beam indices in the next $5$ time spots. Meanwhile, each camera image in ${\bf S}(t)$ could contain multiple objects and thus appear in multiple instances. Hence, the number of the camera images in ${\bf S}(t)$ is only $23916$, much fewer than the instance number. 

%In machine learning, the dataset split to training and validation subsets is important for building and testing a model. It is also well understood that training and validation datasets having a similar distribution benefits the model generality. 
A closer observation of the Viwi dataset reveals that almost all the images in $\mathcal{D'}_{v}$ also exist in $\mathcal{D'}_{t}$. Specifically, out of the $23916$ images, only $13$ and $2$ images uniquely appear in $\mathcal{D'}_{t}$ and $\mathcal{D'}_{v}$, respectively, and all the other $23901$ images appear in both. %In practice, after the model is trained, the images used for beam tracking, i.e., in the test dataset, should not be seen before even they are taken from the same environment. 
As introduced earlier, to avoid the possibility that a model performs well in $\mathcal{D'}_{v}$ just because it memorizes the images in $\mathcal{D'}_{t}$, we reformulate the datasets as follows:
\begin{eqnarray}
\mathcal{D}_{t}&\!\!\!\!\triangleq\!\!\!\!& \mathcal{D'}_{t}[70251: 210787] \cup \mathcal{D'}_{v}[30141: 90389],\\
\mathcal{D}_{v1}&\!\!\!\!\triangleq\!\!\!\!& \mathcal{D'}_{t}[:70251] \cup \mathcal{D'}_{v}[:30141],\\
\mathcal{D}_{v2}&\!\!\!\!\triangleq\!\!\!\!& \mathcal{D'}_{t}[210787:] \cup \mathcal{D'}_{v}[90389:],
\end{eqnarray}
where $\mathcal{D}[a:b]$ represents the dataset $\mathcal{D}$ in the matrix form from the $a^{th}$ to the $(b-1)^{th}$ rows, and the index count starts from 0 (consistent with Python grammar). It can be verified that no data is wasted and the images included in the three sets are mutually exclusive. Moreover, $\mathcal{D}_{t}$ has $200,784$ instances with $11958$ images, $1,993$ from each of the 6 cameras, respectively; $\mathcal{D}_{v1}$ ($\mathcal{D}_{v2}$) has $100,392$ instances with $5979$ images, $1,993$ from camera 1, 2, 3 (camera 4, 5, 6), respectively. 

To this end, we guarantee that the images in each dataset are exclusive from the others. Thus, one can (1) train a model on $\mathcal{D}_{t}$ and test it on $\mathcal{D}_{v1}\cup \mathcal{D}_{v2}$; or (2) train a model on $\mathcal{D}_{t}\cup \mathcal{D}_{v2}$ and test it on $\mathcal{D}_{v1}$, so that the dataset size is close to that in the Viwi dataset \cite{icc2020}; or (3) train a model on $\mathcal{D}_{t}$ and test it on $\mathcal{D}_{v1}$ and $\mathcal{D}_{v2}$ separately, so as to investigate the prediction difference between cameras 1, 2, 3 and cameras 4, 5, 6. Due to the page limit, we choose the option (1) in this paper. 

%we choose the option (3) in this paper. 

\section{The Proposed Scheme} \label{sec:VABT}

In this section, we propose a deep learning model, extended from the model in \cite{Alrabeiah_yolo} for link blockage prediction. %As illustrated in Fig. \ref{fig:DL_model}, 
The proposed model consists of four components -- beam embedding, objective recognition for feature extraction, feature embedding for further dimension reduction, and sequential model for prediction. Although these components also exist in \cite{Alrabeiah_yolo}, the formulation is not the same. In particular, we utilize a wider range of models to explore the proper use of the images in this project. We introduce these four components individually.

% \begin{figure}[t]
%     \centering
%     \includegraphics[align=t, width=1.0\linewidth]{fig1.png}
%     \vspace{-0.1in}
%     \caption{The model structure in our investigation}
%     \label{fig:DL_model}
% \end{figure}

\subsection{Beam Embedding}

To represent each beam index $f$ at time $t$ with a vector that can be fed into the machine learning models, we can directly utilize its beamforming vector ${\bf f}$ from the codebook $\mathcal{F}$. Similar to \cite{Alrabeiah_icc, Alrabeiah_yolo}, we consider each vector ${\bf f}_q$ in $\mathcal{F}$ is independently draw from a Gaussian distribution with zero mean and the identify matrix as covariance\footnote{We also consider $\mathcal{F}$ as an orthogonal codebook owing to its attributes of suppressing multi-user interference and less power variations in wireless systems. However, based on our experiments, there is no significant difference in the final performance evaluation and comparison.}, where the length of the ${\bf f}_q$ vector is assumed to be $2N$ as an example. Note that this beam embedding process is straightforward as it does not need any more than code generation and assignment.

\subsection{Objective Recognition for Feature Extraction}\label{sec:yolo}

Considering the image dimension is much larger than the beam vector dimension, we are inclined to extract image features. Recall that each image contains a moving user of interest, but which user is of interest is unknown. Thus, we can extract some information related to the user trajectories and thus reduce the noise. With this goal, we apply objective detection. In this paper, we utilize the well known pre-trained model -- You Only Look Once (YOLO) \cite{Redmon_cvpr16, Redmon_cvpr17} -- to extract features for predetermined classes including persons, cars, trucks, etc. To fit it in our problem, we retain the features at the last intermediate layer before the prediction is made. Then, we have three options of the output dimension: $[13,13,255]$, $[26,26,255]$, $[52,52,255]$. For speeding up the computation process, we choose the option with $[13,13,255]$ as the output size\footnote{For the other options with larger dimensions, although the recognition resolution improves, more redundant information could also be involved as noise. Thus, we start with the option with the smallest dimension size.}.

\subsection{Feature Embedding for Further Dimension Reduction}

Since the output from objective detection is with the dimension of $13\times13\times 255$ much larger than the beam embedding size $2N=256$, we tend to further reduce the dimension via feature representation. Several approaches, either supervised or unsupervised learning, can be used for dimension reduction. In this paper, we consider three methods below:
\begin{enumerate}
    \item We use supervised learning by assigning one or multiple beam indices (w.r.t. the users) to each image. Then we build a CNN-based model for multi-label multi-class classification, and obtain the feature embedding from an intermediate layer. Here, ``multi-label" captures the multiple users appearing in the same image, and ``multi-class" is because of $N$ beams available to the users. 
    \item Since the dataset distribution is highly skewed in labels and classes, we consider unsupervised learning -- AutoEncoder (AE) -- to represent each image at the bottleneck. To deal with the heavier computation load of AE, in practice the model can be trained offline, as long as the environment does not significantly change.
    \item We consider PCA, a linear and fast-computational method for exploration and comparison. Since feature embedding is only an intermediate component of the entire model and it is unclear what kind of image processing is more favorable for the end-to-end performance, we usually prefer a simpler scheme. 
\end{enumerate}
For the three methods introduced above, we set the embedding dimension as $|{\bf f}_q|$ for uni-GRU and $|{\bf f}_q/4|$ for bi-GRU, the models that we use in the next component. Generally, one can adjust the value so that better algorithms might be discovered. 

\subsection{Sequential Models for Prediction}

%CNN networks are not good at studying the dependencies among data samples in nature, so they are not used to learn the features relation in this paper. In the proposed algorithm, RNN is utilized to perform the future beam prediction based on the learned relation. The recurrent component uses Gated Recurrent Units (GRU) separated by a dropout layer, being followed by a fully connected layer as a classifier. It generates codebook size possibilities and the index of the element with the highest probability in every vector are the predicted beams. 

Note that the processing components introduced above apply for each image and beam only. Consider the observational sequence defined in Sec. \ref{sec:sys} contains $\tau$ time spots, we apply uni-GRU and bi-GRU models for beam prediction. 

Another interesting question is how to incorporate the beam and image embedding together efficiently. In this paper, we consider both alternatively staggering them and concatenating them, which result in $2\tau \times |{\bf f}_q|=16\times 256$-dimensional input to the uni-GRU and $\tau \times (|{\bf f}_q| + |{\bf f}_q/4|) = 8\times 320$-dimensional input to the bi-GRU, respectively. We use $|{\bf f}_q/4|$ rather than $|{\bf f}_q|$ as the size of the AE's bottleneck for achieving better performance, and it can be better tuned as well.

\section{Experimental Results}\label{sec:modif_ret}

In this section, we experiment on the proposed scheme introduced in Sec. \ref{sec:VABT}. Besides the model parameters for each component, the general parameters for training models include: optimizer: Adam, learning rate: 0.001, batch size: 1000, GRU layers: 4, GRU hidden layer dimension: 256, epoch number: 12 (for bi-GRU) and 50 (for bi-GRU), dropout: 0.2, and the loss function: cross entropy loss. 

We start with the baselines by using beams only and then focus on improving the performance with using the images. 

\subsection{Results of the Baseline Solutions}

Four baselines by using beam sequence only are considered: (1) \emph{The modified baseline in \cite{Alrabeiah_icc}}: we consider the baseline introduced in \cite{Alrabeiah_icc} by changing the uni-GRU layers from 1 to 4 and increasing the hidden layer dimension from 20 to 256 for achieving better performance. (2) \emph{The last-step repetition baseline}: for each user instance, we repeat the last-step index in the observational sequence as the next $m$ predictions. This is very simple, with short memory, and can be directly performed on the validation dataset. (3) \emph{The linear regression baseline}: we use linear regression to fit the beam indices in the observational sequence and then apply it for prediction. (4) \emph{The statistical baseline}: based on all data in the first $\tau=8$ columns of the training dataset, we calculate the beam index distribution, from which we randomly draw a beam index for each user instance in the validation as the next $m$ predictions. 

The results of the baselines above are shown in the first four rows of Table \ref{tab:modified_base}. It can be seen that the modified baseline in \cite{Alrabeiah_icc} generally performs the best due to the use of deep learning, and both it and the last-step repetition baseline perform much better than the other two. A little surprisingly, the simple last-step repetition performs even $0.004$ better than the modified baseline in \cite{Alrabeiah_icc} and with much cheaper computation, and the advantage is even more significant, as large as $0.024$, in Score$_5$. This is probably because the user moves with a relatively low speed and/or in the LOS environment, and the beam index also changes slowly. If the user moves in a relatively high speed or in NLOS, then as Score$_5$ suggests, deep learning might not be a good choice for the beam only.

\begin{table}
    \centering
    \begin{tabular}{c|cccc}
    \hline\hline
        Models & Score$_1$ & Score$_3$ & Score$_5$ & TotalScore\\
    \hline
        modified baseline in \cite{Alrabeiah_icc} & {\bf 0.862} & 0.642 & 0.517 & 0.597\\
        Last-step repetition & 0.797 & 0.635 & 0.541 & 0.601\\
        Linear regression & 0.358 & 0.324 & 0.298 & 0.313 \\
        Statistical baseline & 0.039 & 0.035 & 0.033 & 0.034\\
        YOLO + PCA (256) & 0.857 & 0.638 & 0.517 & 0.595\\
        YOLO + CNN (256) & {\bf 0.862} & 0.660 & 0.552 & 0.622\\
        YOLO + AE (256) & {\bf 0.862} & 0.660 & {\bf 0.558} & {\bf 0.626}\\
        YOLO + AE (128) & 0.861 & {\bf 0.662} & 0.552 & 0.623\\
        YOLO + AE (64) & 0.860 & 0.660 & 0.548 & 0.620\\
        %VGG16 + PCA (256) & 0.857 & 0.639 & 0.522 & 0.598\\
        %VGG16 + CNN (256) & 0.862 & 0.66 & 0.552 & 0.622\\
        %VGG16 + AE (256) & 0.862 & 0.651 & 0.534 & 0.609 \\
        %VGG16 + AE (64) & 0.862 & 0.656 & 0.545 & 0.617 \\
        %VGG16 + AE (32) & 0.862 & 0.653 & 0.55 & 0.619\\
    \hline\hline
    \end{tabular}
    \vspace{-0.05in}
    \caption{Performance of a variety of image feature extraction and embedding methods with ``uni-GRU beams + images" based on the modified dataset}\label{tab:modified_base}
    \vspace{-0.15in}
\end{table}

\subsection{Results of the Proposed Scheme by using uni-GRU}

We apply the uni-GRU as the sequential model. For brevity, we use ``model for feature extraction + model for feature embedding (the output dimension)" to represent the corresponding method. For example, ``YOLO + PCA (256)" means we apply YOLO to extract images features and PCA to further reduce the feature dimension, and the output dimension is 256. The results are shown in the last five rows of Table \ref{tab:modified_base}.

Firstly, we compare the performance between PCA, AE, and CNN for feature embedding (see the $5^{th}-7^{th}$ rows of Table \ref{tab:modified_base}). First, AE outperforms the PCA, e.g., the overall score of ``YOLO + PCA (256)" is 0.595, which is 0.027 lower than ``YOLO + AE (256)". This is because AE can well learn non-linear features and thus retains more information than PCA. Second, AE outperforms the CNN, e.g., the overall score of ``YOLO + CNN (256)" is 0.622, which is 0.004 lower than that of ``YOLO + AE (256)". Thus, the combination of YOLO and AE turns out to be the best option in our experiment.

Next, we investigate the performances with varying the embedding dimension. Consider the ``YOLO + AE" as an example (see the $7^{th}-9^{th}$ rows of Table \ref{tab:modified_base}). When we change the output dimension from 256 to 128 and 64, the overall score decreases from 0.626 to 0.623 and 0.620, which are dominated by Score$_5$. This is again consistent with our expectation, because the features with larger dimensions can retain more information and benefit the larger multi-step ahead prediction.

Finally, we apply the same model on the original ViWi dataset \cite{icc2020} and obtained the four scores $0.867$, $0.6894$, $0.588$, $0.653$ like in Table \ref{tab:modified_base}, respectively. The comparison reveals that the experimental performance on our reformulated dataset in Sec. \ref{sec:dataset} significantly drops. For example, for the ``YOLO + AE (256)", the overall score for the modified dataset drops from 0.653 to 0.626, which is consistent with our expectation. %, as the image feature embedding model needs to be generalized to new images in the modified validation, which have never been seen in the modified training dataset.

\subsection{Results of the Proposed Scheme by using bi-GRU}\label{sec:bi-gru}

We also play the bi-GRU model for the beam prediction. The results are shown in the first two rows of Table \ref{tab:memory}. The overall score is $0.687$, which is $0.061$ higher than the best using uni-GRU in Table \ref{tab:modified_base}. However, we also note that the score $0.687$ is achieved regardless of the use of the image. Thus, we can conclude the performance improvement does \emph{not} come from using the image. In fact, for the use of the image sequence, while it helps the beam prediction by providing more useful underlying features behind the environment, it may also neutralize the additional benefits brought by the use of the image. Thus, with the use of bi-GRU, it is of interest whether the image is useful for beam prediction. We will address this interesting and non-straightforward question in Sec. \ref{sec:LOS_NLOS}.  

\begin{table}
    \centering
    \begin{tabular}{c|cccc}
    \hline\hline
        Observation sequence & Score$_1$ & Score$_3$ & Score$_5$ & TotalScore\\
    \hline
        beam only with $\tau = 8$ & 0.856 & {\bf 0.717} & {\bf 0.635} & {\bf 0.687}\\
        beam + image with $\tau = 8$ & {\bf 0.857} & {\bf 0.717} & {\bf 0.635} & {\bf 0.687}\\
        beam + image with $\tau = 6$ & 0.854 & 0.714 & 0.625 & 0.680\\
        beam + image with $\tau = 4$ & 0.846 & 0.700 & 0.605 & 0.663\\
    \hline\hline
    \end{tabular}
    \vspace{-0.05in}
    \caption{Performance w.r.t. the observation sequence length using the model of ``bi-GRU beams + images"}
    \label{tab:memory}
    \vspace{-0.15in}
\end{table}

\subsection{Trade-off between Memory Size $\tau$ and Performance}

%So far all the experiments except for the last-step repetition utilize the observational sequence with length $\tau=8$. 
Usually, the larger $\tau$, the better (or no worse) prediction. However, a larger $\tau$ also implies the need for higher computation and larger memory. Thus, one may ask what a reasonable value of $\tau$ achieves a certain performance. By varying the value of $\tau$ (and changing the input size of the model), we obtain the prediction results in the last three rows of Table \ref{tab:memory}, indicating the longer observational sequence indeed benefits the prediction. For example, with $\tau=4$ the model achieves $0.605$ for $m=5$, and can be readily improved to $0.625$ and $0.635$ for the settings with $\tau=6,~8$, respectively, which also implies marginal diminishing return. To achieve the target performance, we expect $\tau$ as smaller as possible, e.g., $\tau=6$ suffices to achieve 0.85 as the target score for $m=1$. Using $\tau=8$ can only bring marginal improvement, and requires much heavier computational load and $33\%$ more storage.

\section{Does the Image Really Assist Beam Prediction?}\label{sec:LOS_NLOS}

In this section, we follow the question in Sec. \ref{sec:bi-gru} and try to shed light on how to properly use the image in addition to the beam for beam prediction with bi-GRU. In addition, on the ICC competition website, it is interesting that the highest score of the participants is only 0.38433, even much lower than the modified baseline and the last-step prediction by using beam only shown in Table \ref{tab:modified_base}. We suspect that the unseen test dataset might contain many more moving users in NLOS areas, because the user in NLOS would see more unstable beam over time than in LOS. Since the ViWi dataset does not tell whether each user instance is in LOS or NLOS area, we propose to use the standard deviation (std) of the beam index sequence to indicate the sequence stability, i.e., the larger std, the more unstable, and thus more likely to be in NLOS. 

With two predetermined std thresholds to form three cluster intervals, representing LOS-like, light NLOS-like, and serious NLOS-like, we assign each user instance to one cluster. As an example, we use $\mathcal{D}_{t}$ for training and $\mathcal{D}_{v1}[:5000]$ for validation\footnote{This assumption is made to investigate if the training and validation datasets contain quite different ratio between the numebrs of LOS/NLOS-like data instances. In particular, the std of the beam sequences in $\mathcal{D}_{t}$ is with mean 4.922 and median 0.484, but with 7.842 and median 1.378 for $\mathcal{D}_{v1}[:5000]$. Moreover, the percentage of the user instances with constant beam indices, i.e,, std$=0$, reduces from $38.74\%$ for $\mathcal{D}_{t}$ to only $20.6\%$ for $\mathcal{D}_{v1}[:5000]$, nearly $50\%$ off. Hence, we conclude the significant decrease in score is due to the fact that more links are NLOS in testing than in training.}. By choosing ``0" and ``2" as the std thresholds, each dataset can be split into three subsets. For brevity, we name each subset with a corresponding letter as shown in Table \ref{tab:representation}. The results are shown in Table \ref{tab:NLOS} for Score$_5$ only.

\begin{table}
    \centering
    \begin{tabular}{c|ccc}
    \hline\hline
        std range &  std = 0 & 0 $<$ std $\leq$ 2 & std $>$ 2 \\
    \hline
        Training $\mathcal{D}_{t}$  & $\mathcal{A}_t$ & $\mathcal{B}_t$ & $\mathcal{C}_t$\\
        Validation $\mathcal{D}_{v1}[:5000]$ & $\mathcal{A}_v$ & $\mathcal{B}_v$ & $\mathcal{C}_v$\\
    \hline\hline
    \end{tabular}
    \vspace{-0.05in}
    \caption{The names of subsets w.r.t. the std ranges}
    \label{tab:representation}
    \vspace{-0.1in}
\end{table}

\begin{table}
    \centering
    \begin{tabular}{c|c|cc}
    \hline\hline
        Training & Validation &  beams only & beams + images\\
    \hline
        $\mathcal{A}_t$ & $\mathcal{A}_v$ & {\bf 0.593} & {\bf 0.595} \\
        $\mathcal{A}_t\cup\mathcal{B}_t\cup\mathcal{C}_t=\mathcal{D}_t$ & $\mathcal{A}_v$ & 0.583 & 0.585 \\
        \hline
        $\mathcal{B}_t$ & $\mathcal{B}_v$ & 0.471 & 0.474 \\
        $\mathcal{A}_t\cup\mathcal{B}_t$ & $\mathcal{B}_v$ & {\bf 0.480} & {\bf 0.481} \\
        $\mathcal{A}_t\cup\mathcal{B}_t\cup\mathcal{C}_t=\mathcal{D}_t$ & $\mathcal{B}_v$ & 0.467 & 0.470 \\
        \hline
        $\mathcal{C}_t$ & $\mathcal{C}_v$ & 0.275 & 0.279 \\
        $\mathcal{B}_t\cup\mathcal{C}_t$ & $\mathcal{C}_v$ & {\bf 0.279} & {\bf 0.290} \\
        $\mathcal{A}_t\cup\mathcal{B}_t\cup\mathcal{C}_t=\mathcal{D}_t$ & $\mathcal{C}_v$ & 0.279 & 0.280 \\
    \hline\hline
    \end{tabular}
    \vspace{-0.05in}
    \caption{Score$_5$ performances of the model using ``bi-GRU beams + images" under different link situation}
    \label{tab:NLOS}
    \vspace{-0.15in}
\end{table}

%For brevity, we use the name of ``training group name/validation group name" to represent corresponding experiments. For example, ``B+C/c" means we use the instances in $\mathcal{D}_{t}$ with observed beam sequence std $\textgreater$ 0 as training group and use the instances in $\mathcal{D}_{v1}[:5000]$ with observed beam sequence std $\textgreater$ 2 as validation group, then perform the bi-GRU model to predict the next $m=5$ beams. 
Firstly, let us look at each individual validation subset: %Firstly, we compare the result from the same validation group by training different instances.
(1) for the LOS-like case, we train two models on $\mathcal{A}_t$ and $\mathcal{D}_t$ and validate them on $\mathcal{A}_v$, respectively (see the first two rows of Table \ref{tab:NLOS}). It can be seen that including NLOS-like data to train the model actually hurts the model performance in validation. In addition, no matter for which model, using the image brings extra $0.002$ improvement in the score. 
(2) for the light NLOS-like case, we train three models on three datasets and validate them on $\mathcal{B}_v$, respectively (see the middle three rows of Table \ref{tab:NLOS}). Compared to training on $\mathcal{B}_t$ only, interestingly, it implies including LOS-like data $\mathcal{A}_t$ significantly help the model in validation, but including serious NLOS-like data $\mathcal{C}_t$ significantly hurts the model in validation.
(3) for the serious NLOS-like case, we train three models on three datasets and validate them on $\mathcal{C}_v$, respectively (see the last three rows of Table \ref{tab:NLOS}). Compared to training on $\mathcal{C}_t$ only, interestingly, it shows including light NLOS-like data $\mathcal{B}_t$ significantly help the model in validation, but further including LOS-like data $\mathcal{A}_t$ surprisingly neutralizes the benefits. While for beam only, the improvement can achieve $0.004$, but when the image come to play, there is no significant improvement anymore. 

Second, from the LOS-like to the light NLOS-like and then to the serious NLOS-like, the scores of ``beam only" and ``beam $+$ image" both significantly drop, suggesting that NLOS is the main obstacle. Moreover, the score improvement owing to the use of the image turn out to be $0.002$, $0.001$, $0.011$, respectively. While the first two numbers imply the improvement is very marginal, the last number indicates the image use can significantly improve the beam prediction for the user in serious NLOS environment, and the non-straightforward solution is to include the light NLOS-like data but not the LOS-like as extra data augmentation to train the model.

Finally, adding up the scores of the three clusters weighted by the cardinality of each cluster results in TotalScore $=0.426$, significantly higher than $0.414$ obatined by training on $\mathcal{D}_t$ and validation on $\mathcal{D}_v$. This suggests incorporating data clustering helps beam prediction of the user in serious NLOS-like cases.

To conclude, splitting the user instances according their link LOS/NLOS status can improve the prediction performance especially in the NLOS, and the key is to find an appropriate subset of the training dataset for training a model. Note that here we consider only three clusters, and many insights are already non-trivial. If we consider more clusters, %based on std or some other metrics that better characterizes the extent of LOS/NLOS, the combination of subsets to form the training data would be more complicated, and 
it is much harder to design a systematic approach to find the optimal combination of the corresponding subsets for training a model.

\section{Conclusion}

For vision-aided beam tracking, we developed a deep learning approach via beam embedding, image recognition, image feature embedding, and sequential models, and the experimental results suggest the overall score $0.687$ can be achieved in a variation of ViWi dataset in \cite{icc2020}. We also explore the proper use of the image for better beam prediction than using the beam only. Our experiments demonstrate that using the image indeed improves the beam prediction particularly when the user is in NLOS-like environment, and the key is to carefully design the dataset with clustering for training a model. Future work could include improving the beam prediction for the user in NLOS via developing better models and obtaining higher-quality data that can better reflect the use cases, and interpreting the performance score with real system metrics.

%% the bibliography file.
% \bibliographystyle{ieeetr}
% \bibliography{ref}

\begin{thebibliography}{99}

\bibitem{icc2020} 
The official website of the vision-aided beam tracking data competition at IEEE ICC 2020: `` \url{https://viwi-dataset.net/viwi-bt.html}."

\bibitem{Alrabeiah_yolo} 
M. Alrabeiah and A. Alkhateeb, ``Deep learning for mmwave beam and blockage prediction using sub-6 ghz channels," {\em IEEE Transactions on Communications,} vol. 68, no. 9, pp. 5504–5518, 2020.

\bibitem{Alrabeiah_icc} 
M. Alrabeiah, J. Booth, A. Hredzak, and A. Alkhateeb, ``Viwi visionaided mmWave beam tracking: Dataset, task, and baseline solutions," {\em arXiv preprint arXiv:2002.02445}, 2020.

\bibitem{Alrabeiah_resnet}
M. Alrabeiah, A. Hredzak, and A. Alkhateeb, ``Millimeter Wave Base Stations with Cameras: Vision Aided Beam and Blockage Prediction," {\em arXiv:1911.06255v2 [cs.IT]}, Nov 2019.

\bibitem{master_thesis} 
Shrreenithi Srinivasan, ``Beam Prediction Using Deep Learning Methods for mmWave Communications," {\em \url{https://csus-dspace.calstate.edu/bitstream/handle/10211.3/217943/Shrreenith_Srinivasan_Report.pdf}}, 2020.

\bibitem{Alouini} 
Y. Tian, G. Pan, and M. Alouini, ``Applying Deep-Learning-Based Computer Vision to Wireless Communications: Methodologies, Opportunities, and Challenges," {\em arXiv:2006.05782v4,} Dec 2020.

\bibitem{Redmon_cvpr16} 
J. Redmon, S. Divvala, R. Girshick, and A. Farhadi, ``You only look once: Unified, real-time object detection," {\em in Proc. of the IEEE conference on computer vision and pattern recognition,} 2016, pp. 779–788.

\bibitem{Redmon_cvpr17} 
J. Redmon and A. Farhadi, ``YOLO9000: better, faster, stronger," {\em in Proc. of the IEEE CVPR 2017,} 2017, pp. 7263–7271.

% \bibitem{Lin_ms} 
% T.-Y. Lin, M. Maire, S. Belongie, J. Hays, P. Perona, D. Ramanan, P. Dollar, and C. L. Zitnick, ``Microsoft coco: Common objects in context," {\em in European conference on computer vision,} Springer, 2014, pp. 740–755.


\end{thebibliography}

% \vspace{12pt}
% \color{red}
% IEEE conference templates contain guidance text for composing and formatting conference papers. Please ensure that all template text is removed from your conference paper prior to submission to the conference. Failure to remove the template text from your paper may result in your paper not being published.

\end{document}